%% file: acl2019.tex
\documentclass[11pt,a4paper]{article}
\usepackage{authblk}
\usepackage[hyperref]{acl2019}
\usepackage{times}
\usepackage{latexsym}
\usepackage{times}
\usepackage{latexsym}
\usepackage{graphicx}
\usepackage{makecell}
\usepackage{pbox}
\usepackage{caption}
\usepackage{subcaption}
\usepackage{enumitem}

\usepackage{hhline}
\usepackage{comment}

\newcommand{\cmmnt}[1]{\ignorespaces}

\usepackage{lipsum} 
\usepackage{enumitem}
\usepackage{amsmath}

\usepackage{url}
\usepackage{url}

\aclfinalcopy 

\title{Investigating Evaluation of Open-Domain Dialogue Systems With Human Generated Multiple References}

\author[1]{Prakhar Gupta}
\author[1]{Shikib Mehri}
\author[1]{Tiancheng Zhao}
\author[2]{\authorcr Amy Pavel}
\author[1]{Maxine Eskenazi}
\author[1,2]{Jeffrey P. Bigham}
\affil[1]{Language Technologies Institute\\
Carnegie Mellon University\\}
\affil[2]{Human-Computer Interaction Institute\\
Carnegie Mellon University\\}
\affil[ ]{\texttt {\{prakharg,amehri,tianchez,apavel,max+,jbigham\}@cs.cmu.edu}}

\date{}

\begin{document}
\maketitle
\begin{abstract}
The aim of this paper is to mitigate the shortcomings of automatic evaluation of open-domain dialog systems through multi-reference evaluation. Existing metrics have been shown to correlate poorly with human judgement, particularly in open-domain dialog. One alternative is to collect human annotations for evaluation, which can be expensive and time consuming. To demonstrate the effectiveness of multi-reference evaluation, we augment the test set of DailyDialog with multiple references. 
A series of experiments show that the use of multiple references results in improved correlation between several automatic metrics and human judgement for both the quality and the diversity of system output.
\end{abstract}
\input{introduction.tex}

\input{relatedworks.tex}

\input{methodologyt.tex}

\input{datacollection.tex}

\input{experiments.tex}

\input{acknowledgements.tex}

\bibliography{references}
\bibliographystyle{acl_natbib}

\appendix
\clearpage
\section{Further Notes on Data Collection Experiments}
\label{sec:supplemental}
\begin{figure*}[tb]
    \centering
    \includegraphics[width=0.9\textwidth]{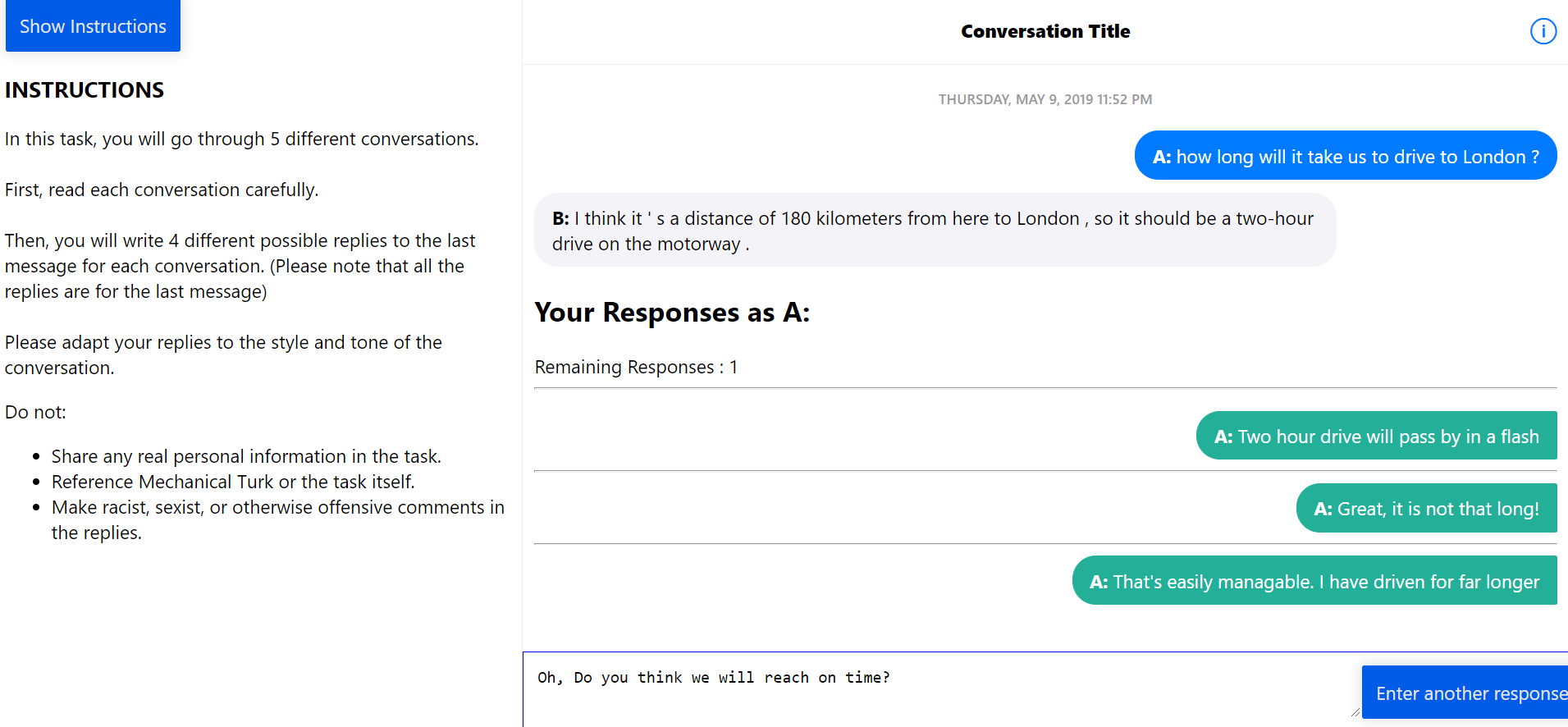}
    \caption{Interface used for multi-reference data collection.}
    \label{fig:interface1}
\end{figure*}

The interface designed for multi-reference data collection is shown in Figure \ref{fig:interface1}. The final design of the interface incorporates improvements based on multiple rounds of experiments and interviews on a small set of users. The workers were shown a modal box with instructions and several good and bad examples before they start the task. Then they are shown 5 contexts for a HIT, one by one. For each context, they are asked to write 4 diverse responses in the Textbox provided. Workers can enter multi-line responses and submit a response by pressing enter or clicking on a button. They are shown the number of remaining responses they need to enter for the conversation. We also record the timestamps for click and enter presses in the interface. We prevent workers from entering replies shorter than 2 characters, the exact same reply more than 1 time and show them a warning prompt if enter their response too quickly consistently. 

\begin{table}[b]
\centering
\begin{tabular}{|l|l|l|l|}
\hline
\textbf{Metric} & \textbf{4R1W} & \textbf{2R2W} & \textbf{1R4W} \\ \hline \hline
SelfBLEU-1      & 0.3809                          & 0.3662                           & 0.4403                          \\ \hline
SelfBLEU-2      & 0.1778                          & 0.1618                           & 0.2657                          \\ \hline
SelfBLEU-3      & 0.0955                          & 0.0851                           & 0.2045                          \\ \hline
SelfBLEU-4      & 0.0548                          & 0.0449                           & 0.1748                          \\ \hline
Distinct-1      & 0.7266                          & 0.7522                           & 0.7082                          \\ \hline
Distinct-2      & 0.9240                          & 0.9346                           & 0.8782                          \\ \hline
Distinct-3      & 0.9621                          & 0.9692                           & 0.9092                          \\ \hline
Gt-BLEU-1       & 0.1213                          & 0.1165                           & 0.1296                          \\ \hline
Gt-BLEU-2       & 0.0258                          & 0.0259                           & 0.0352                          \\ \hline
Gt-BLEU-3       & 0.0091                          & 0.0111                           & 0.0136                          \\ \hline
Gt-BLEU-4       & 0.0033                          & 0.0032                           & 0.0033                          \\ \hline
\end{tabular}
\caption{Diversity and relevance for different modes of data collection.}
\label{table:style}
\end{table}

\textbf{Data Collection modes} - 
For the collection of 4 responses per context, we have the following options - A) 4R1W- Collect 4 responses from a single worker B) 2R2W- Collect 2 responses each from 2 separate workers, and C) 1R4W - Collect 1 response each from 4 separate workers. In order to decide between these collection modes, we designed an experiment where, for 100 random contexts, we collected 4 responses using all three styles A), B) and C). In order to decide the best option, we measured lexical diversity across the 4 responses using self-BLEU \cite{Zhu:2018:TBP:3209978.3210080texygen} and Distinct \cite{li-etal-2016-diversity-distinct} metrics, and the collected responses' relevance through the average BLEU score of the multi-reference responses with the ground truth (Gt-BLEU) in the dataset. The results are reported in Table \ref{table:style}. 

To calculate Self-BLEU, we calculate the BLEU score for every response by treating the response as a hypothesis and the others as the references, and we define the average BLEU scores calculated this way to be the Self-BLEU of the response set. A higher Self-BLEU score implies less diversity in the set. We observe that 4R1W and 2R2W achieve higher lexical diversity than 1R4W. This is because when a worker is asked to write multiple responses, they can make their responses more diverse conditioned on their previous responses. Relevance metrics Gt-BLEU-1,2,3,4 indicate that 1R4W achieve higher lexical similarity with the ground truth response in the dataset, followed by 4R1W. We chose the 4R1W mode, that is, a collection of 4 responses from 1 worker, to balance the diversity and relevance metrics. 

\textbf{Instructions for annotation collection for Diversity Study}

We provided following instructions to the workers for collecting diversity ratings-
``Please read the following conversation between two persons. Then read some possible follow-up responses for the conversation. You will be shown 5 sets of responses, with 5 responses in each set. 
For each response set, first select the responses you think are appropriate responses for the conversation. Then use the sliders to rate the diversity of the response set, that is, how many of the appropriate responses in the response set had different meanings or were different replies. Please provide the diversity score only for the appropriate responses you have marked. The diversity score should not be more than the number of appropriate responses in that set.''
These instructions were followed by an example to make the task clear.

\section{Choice of dataset}

There are only a few open-domain multi-reference datasets and they have been collected artificially either by retrieval \cite{xu-etal-2018-lsdscc, galley-etal-2015-deltableu} or are very small in scale \cite{Sugiyama2019automaticevaluation}. 
Therefore we augmented the original test set of the DailyDialog dataset \cite{li-etal-2017-dailydialog}, which has a sufficiently large test set. Conversations in DailyDialog cover 10 different topics on daily life. 
We chose to augment the DailyDialog dataset due to the following reasons- 
1) The dialogs in this dataset are about daily conversation topics and thus it is easier to augment them using crowdsourcing.
2) The dialogs in this dataset are generally more formal than datasets such as the Twitter Dialog Corpus \cite{ritter2011data-twitter} and Ubuntu Corpus \cite{lowe-etal-2015-ubuntu} which contain noise such as typos and slangs.
3) The dialogs generally have a reasonable number of turns, which makes it easier for a person to understand the context and generate a reply. 
Therefore, given the size of the original DailyDialog test set and the above-mentioned properties of the dataset, we chose to augment the test set of DailyDialog.

\begin{table}[tb]
\begin{tabular}{|l|c|c|}
\hline
\textbf{Reference} & \multicolumn{1}{l|}{\textbf{Original}} & \multicolumn{1}{l|}{\textbf{Multi-reference}} \\ \hline
Unique 1-gram      & 17.55                                  & 23.62                                         \\ \hline
Unique 2-gram      & 27.88                                  & 58.69                                         \\ \hline
Unique 3-gram      & 21.79                                  & 50.34                                         \\ \hline
\end{tabular}
    \caption{Comparison of number of unique n-grams in original versus multiple references. }
    \label{table:uniquengram}
\end{table}

\vspace{2mm}
\noindent
\textbf{Dataset quality continued}\\
We present the average number of unique 1, 2 and 3 grams in the original ground truth and the set of collected multi-reference ground truth in Table \ref{table:uniquengram}. The higher number of unique ngrams in the multi-reference ground truth indicates that the new ground truth captures more variation in the set of possible responses.
\end{document}

%% file: introduction.tex
\section{Introduction}

%
%
%
%
Dialogue agents trained end-to-end to hold open-domain conversations have recently progressed rapidly, generating substantial interest ~\cite{ghazvininejad2018knowledge, serban2017hierarchical, Serban:2016:BED:3016387.3016435,Sordoni2015ANN,vinyals2015neural}.
Development of these systems is driven by available data and benchmarks based on only a single ground truth reference response for a given context.
%
However, such single-reference evaluation does not account for all the plausible responses for any given conversational context (Table~\ref{table:multirefeg}). This is known as the \textit{one-to-many} response problem \citep{zhao-etal-2017-learning-cvae}.
Computing word-overlap metrics against a single-reference response may penalize perfectly valid responses ~\cite{deriu2019survey} (e.g., ``Was anything stolen?'', ``Is anyone hurt'') that deviate from the particular target response (``When was the break-in?''). Unlike human evaluation, 
automatic evaluation with a single-reference may also disproportionately benefit models that produce generic responses with more probable words (e.g., ``I don't know'') which is known as the dull-response problem \citep{li2016deep}. As a result, single-reference evaluations 
correlate weakly with human judgments of quality \cite{liu-etal-2016-howtonotevaluate}.




\begin{table}[]
    \centering
    \begin{tabular}{l}
    \hline
\textbf{Dialogue Context:}\\
\textit{Person A:} 911 emergency. What is the\\ problem?\\
\textit{Person B:} I would like to report a break-in.\\
\hline 
\textbf{Single-reference Response:} \\
When was this break-in?\\
\hline
\textbf{Other Valid Responses:}\\
Was anything stolen?\\
Is anyone hurt or injured?\\
Is the perpetrator still inside the house?\\
I will send someone right away.\\
\hline
\end{tabular}
\caption{Example of a dialogue context where appropriate responses do not share words and meaning with a single-reference response.} 
\label{table:multirefeg}
\end{table}

To address these problems, this paper proposes to carry out automatic evaluation using multiple reference responses instead of a single-reference. Multiple reference evaluation is attractive for several reasons. First, the additional information in the multiple reference response can be used to provide more robust quality evaluation under the one-to-many condition. Second, we can use the multiple references to better measure the diversity of the model, which is a widely studied topic in open-domain response generation~\cite{kulikov2018importance-beam, li-etal-2016-diversity-distinct, Zhang:2018:GID:3326943.3327110diverseadv,li-etal-2016-persona,zhao-etal-2017-learning-cvae, gao2019jointly}.

Prior explorations in this area either rely on synthetically created or small scale reference sets~\cite{galley-etal-2015-deltableu,qin-specia-2015-truly}, or perform experiments only on a small set of metrics focused on only response quality~\cite{Sugiyama2019automaticevaluation}.
Our investigations for using multiple references for automatic evaluation covers the following aspects - 1) We propose methodology for evaluating both the quality and the diversity of generated responses using multiple references. 2) The proposed evaluation framework is metric-agnostic and the experiments cover a large spectrum of existing metrics, and 3) We augmented the exiting test set of DailyDialog dataset~\cite{li-etal-2017-dailydialog} with multiple references and perform human judgment correlation studies with human-generated references. 
Our extensive experimental results show that using multiple test references leads to significantly better correlation of automated metrics with human judgment in terms of both response quality and diversity. This suggests that the use of multiple references serves to make automatic metrics more reliable mechanisms for evaluating open-domain dialogue systems. 
Moreover, follow up studies are conducted to better understand the nature of the multi-reference evaluation, such as the number of reference responses needed to achieve high correlation.

The contributions of this paper are:
\begin{enumerate}[noitemsep,topsep=0pt,parsep=0pt,partopsep=0pt]
    \item We show that multi-reference evaluation achieves better correlation with human judgments both in quality and in diversity. 

\item We analyze the effect of varying the number of reference responses on the correlation with human quality judgements.
\item We construct and release an open-domain multi-reference test dataset\footnote{https://github.com/prakharguptaz/multirefeval}.
\end{enumerate}

%% file: relatedworks.tex
\section{Related work}
The need for reliable and consistent automatic evaluation methodologies has lead to increasing interest in dialogue system evaluation in recent years.
In domains such as machine translation and captioning, n-gram overlap metrics such as BLEU~\cite{Papineni:2002:BMA:1073083.1073135BLEU}, ROUGE~\cite{lin-2004-rouge} and METEOR~\cite{Lavie:2007:MAM:1626355.1626389METEOR} 
correlate well with human judgement. Several embedding-based metrics have been proposed as well, including Greedy Matching~\cite{rus2012comparison-greedymatching} and Vector Extrema~\cite{forgues2014bootstrapping-extrema}. These automatic metrics, however, do not generalize well to open-domain dialogue due to the wide spectrum of correct responses, commonly known as the one-to-many problem \citep{zhao2017learning}.
Recent work has proposed several trainable evaluation metrics to address this issue. RUBER \cite{tao2018ruber} evaluates generated responses based on their similarity with the reference responses and their relatedness to the dialogue contexts. \citet{lowe-etal-2017-towards-ADEM} trained a hierarchical neural network model called ADEM to predict the appropriateness score of responses. However, ADEM requires human quality annotation for training, which is costly. \citet{sai2019reADEMre} recently showed that trainable metrics are prone to gamification through adversarial attacks. While past work has focused on inventing new metrics, this paper instead aims to demonstrate that the correlation of existing metrics can be improved through the use of multiple references for evaluation in open-domain settings.


Prior attempts leveraged multiple references to improve evaluation in the context of text generation. \citet{qin-specia-2015-truly} proposed variants of BLEU for machine translation based on n-gram weighting. In the dialogue domain, \citet{galley-etal-2015-deltableu} proposed Discriminative BLEU, which leverages several synthetically created references obtained with a retrieval model from Twitter corpus. 
\citet{Sordoni2015ANN} also followed a similar retrieval procedure for multiple-reference evaluation. Since both of them created their reference sets through retrieval followed by a rating step, their multi-reference sets do not reflect the natural variability in responses possible for a context.  \citet{Sugiyama2019automaticevaluation} proposed a regression-based evaluation metric based on multiple references. The small set of metrics and few test sentences shows promise, but also the need for further exploration. We go further with a comparison of single and multiple references for response quality evaluation and an examination of multiple references for diversity evaluation. 
This paper is the first, to our knowledge, to create a large test set of several human-generated references for each context. We believe that it is also the first to perform human correlation studies on a variety of automatic metrics for both quality and diversity.

Evaluating diversity in dialogue model responses has been studied recently. The most commonly used metric is Distinct~\cite{li-etal-2016-diversity-distinct}, which calculates the ratios of unique n-grams in generated responses. Distinct
is, however, computed across contexts and does not measure if a model can generate multiple valid responses for a context.
\citet{xu-etal-2018-lsdscc} proposed Mean Diversity Score (MDS) and Probabilistic Diversity Score (PDS) metrics for diversity evaluation over groups of multiple references over a set of retrieved references.  \citet{hashimoto2019unifyingHUSE} proposed a metric for a unified evaluation of quality and diversity of outputs, which however depends on human judgements. \citet{zhao-etal-2017-learning-cvae} proposed precision/recall metrics calculated using multiple hypotheses and references as an indicator of appropriateness and coverage. In this paper we leverage their recall-based metrics in our multi-reference based evaluation of diversity.



%% file: methodologyt.tex
\section{Methodology}
We evaluated the performance of dialogue response generation models from two aspects: \textbf{quality} and \textbf{diversity}. Quality tests the appropriateness of the generated response with respect to the context, and diversity tests the semantic diversity of the appropriate responses generated by the model.

We first describe the evaluation procedures used for the conventional single-reference setting. Then we present the proposed multi-reference evaluation. 
We define a
generalized metric to be $\operatorname{d}(y, r)$ which takes a produced output $y$ and a reference output $r$, and produces a matching score that measure the level of similarity between $y$ and $r$. We discuss options for $d$ in Table~\ref{table:metrics}. 

\subsection{Baseline: Single-reference Evaluation}
\subsubsection{Quality}
During single-reference evaluation, there is only one reference response $r$. As such, for a given metric $d$, the single-reference score will be $\operatorname{d}(y,r)$.

\subsubsection{Unreferenced Diversity}
Most prior work 
concentrates on unreferenced diversity evaluation since referenced diversity evaluation requires a multi-reference dataset.
Unreferenced evaluation refers to diversity evaluation methods which ignore the reference responses, and instead compute diversity as a function only of the generated responses. 
The Distinct~\cite{li-etal-2016-diversity-distinct} metric calculates diversity by calculating the number of distinct n-grams in generated responses as a fraction of the total generated tokens. This score is calculated at the system level - over the set of responses generated for all the contexts in test set.
Given a set of system responses for the same context, 
Self-BLEU~\cite{Zhu:2018:TBP:3209978.3210080texygen} sequentially treats each one of the generated responses as the hypothesis and the others as references. This score is computed for every context and then averaged over all contexts. A lower Self-BLEU implies greater diversity since system outputs are not similar to one another.

\begin{table*}[!htb]
\centering {\renewcommand{\arraystretch}{1.9}
\small
\centering
\begin{tabular}{|l@{\hskip2pt}|@{\hskip1pt}l@{\hskip1pt}|l|}
\hline
\textbf{Metric}             & \textbf{Reference } & \textbf{Description}  \\ \hline
\multicolumn{3}{|c|}{\textbf{Word-overlap based metrics}} \\ \hline
BLEU                  & \citet{Papineni:2002:BMA:1073083.1073135BLEU}                     & \pbox{20cm}{BLEU is based on n-gram overlap between the candidate and reference \\ sentences. It includes a brevity penalty to penalize short candidates.}                             \\ \hline


METEOR                  & \citet{Lavie:2007:MAM:1626355.1626389METEOR}                   & \pbox{20cm}{The harmonic mean of precision and recall between the candidate and \\ reference based on a set of alignments between the two.}                             \\ \hline

ROUGE-L & \citet{lin-2004-rouge} & \pbox{25cm}{An F-measure based on the Longest Common Subsequence (LCS)\\ between the candidate and reference utterances.} \\ \hline  

\multicolumn{3}{|c|}{\textbf{Embedding based metrics}} \\ \hline

\pbox{25cm}{Embedding \\ Average } &
\citet{wieting2015towards}, others
& \pbox{25cm}{Computes a sentence-level embedding of $r$ and $c$ by averaging the\\ embeddings of the tokens composing the sentences. }\\ \hline 

\pbox{25cm}{Vector \\  Extrema } &
\citet{forgues2014bootstrapping-extrema}             & \pbox{20cm}{Computes a sentence-level embedding by taking the most extreme value of\\ the embeddings of tokens of the sentence for each dimension of the embedding.}                             \\ \hline 

\pbox{25cm}{Greedy \\ Matching } &
\citet{rus2012comparison-greedymatching}               & \pbox{20cm}{Each word in the candidate sentence is greedily matched to a word in the\\ reference sentence based on the cosine similarity of their embeddings.\\ The score is then averaged for each word in the candidate sentence.}                             \\ \hline 

\pbox{25cm}{Skip-Thought} & \citet{kiros2015skipthought} & \pbox{20cm}{Uses a recurrent network to encode a given sentence into a sentence level\\ embedding. We use the pre-trained vectors and implementation provided\\ by \cite{sharma2017relevance-maluuba}.} \\ \hline

{GenSen} & \citet{subramanian2018learning-gensen} & \pbox{20cm}{Generates a sentence level embedding through a sequence-to-sequence model \\ trained on a variety of supervised and unsupervised objectives in a multi-task\\ framework.} \\ \hline 






\end{tabular}
\caption{Metrics used for both quality and diversity evaluation.}
\label{table:metrics}
}
\end{table*}

\subsection{Proposed: Multi-Reference Evaluation}
\subsubsection{Quality}

In multi-reference evaluation, a given context has multiple valid responses $R = \{r_1, r_2, ..., r_n\}$. As such, for a given metric $d$, the multi-reference score can be computed as:
\begin{equation}
\operatorname{score}(y,R)= \max_{r \in R} {\operatorname{d}(y, r)} 
\end{equation}

We score the system output against only the closest reference response because there are multiple diverse and valid responses for a given context.



\subsubsection{Referenced Diversity}
A multi-reference test set also allows referenced diversity evaluation. For a given context $c$, we are given multiple reference responses $R = \{r_1, r_2, ..., r_n\}$ and multiple system outputs $Y = \{y_1, y_2, ..., r_m\}$. For a given metric, $d$, we compute recall~\cite{zhao-etal-2017-learning-cvae}, or \textit{coverage}, as follows:
\vspace{-2mm}
\begin{equation}
    \operatorname{recall}(\mathrm{c})=\frac{\sum_{j=1}^{M} \max _{i \in[1, N]} \operatorname{d}\left(y_{i}, r_{j}\right) )}{M}
\end{equation}

For each of the multiple reference responses, we consider the highest-scoring system output, then average these scores across the reference responses. A system that generates outputs covering a large portion of the reference responses thus receives a higher recall score. 



\subsection{Metrics}

We consider several metrics for quality and diversity evaluation including (1) word-overlap metrics, and (2) embedding-based metrics.
We describe the metrics in Table \ref{table:metrics}. Each metric represents an instantiation of the generalized scoring function $d$.

\subsection{Compared Models}

Our experiments are conducted using four models: a retrieval model and three different generative models. We treat human generated responses as an additional model.

\vspace{1mm}
\noindent
\textbf{Human}: 
To represent ideal model performance for a particular context, we use a human-generated response for that context.

\vspace{1mm}
\noindent
\textbf{Dual Encoder:} A strong baseline for dialogue retrieval is the Dual Encoder (DE) architecture \cite{DBLP:conf/sigdial/LowePSP15ubuntu}. The model first encodes a given dialogue context and response using an LSTM encoder. It then takes the dot-product of the two latent representations to output the likelihood of the response. The Dual Encoder is trained to differentiate between correct responses, and uniformly sampled negative responses. During inference, however, it chooses a correct response for a given context out of all the responses that occur in the training set.

\vspace{1mm}
\noindent
\textbf{Seq2Seq:} Sequence-to-sequence (Seq2Seq) networks \cite{sutskever2014sequence} are a typical baseline for dialogue systems \cite{vinyals2015neural}. Our model consists of an LSTM encoder, an LSTM decoder and an attention mechanism \citep{bahdanau2014neural}.  

\vspace{1mm}
\noindent
\textbf{HRED:} Hierarchical Recurrent Encoder Decoder networks (HRED)  \cite{Serban:hred} are a modification of Seq2Seq networks. Rather than encoding the context as a sequence of words, the encoding of the context is done in a two-step process. First, all the utterances of a context are independently encoded by an LSTM utterance encoder. Second, given the latent representations of each utterance, a context encoder encodes the dialogue context. The attention mechanism of the decoder attends over the timesteps of context encoder.

\vspace{1mm}
\noindent
\textbf{CVAE:} The Conditional Variational Autoencoder (CVAE) model \cite{zhao-etal-2017-learning-cvae}. CVAE models incorporate discourse-level latent variables in HRED, in which the latent variables represent the discourse-level intentions of the system. Specifically, we reproduce the CVAE network from~\cite{zhao-etal-2017-learning-cvae}, where the latent variables follow a multivariate Gaussian distribution with a diagonal covariance matrix. The dimension of the latent variable is 256. To have a fair comparison, the rest of the structure is the same as the HRED with bidirectional LSTM utterance encoders and LSTM context encoder and response decoder. To alleviate the posterior collapse issue for training text CVAEs~\cite{bowman2016generating}, we use bag-of-words auxiliary loss~\cite{zhao-etal-2017-learning-cvae} and KL-annealing~\cite{bowman2016generating}.

%% file: datacollection.tex
\begin{table*}[!htb]
\small
\centering
\begin{tabular}{|c|c|c|c|c|c|}
\hline
\textbf{Reference}             & \textbf{\begin{tabular}[c]{@{}c@{}}Very\\ Appropriate\end{tabular}} & \textbf{Appropriate} & \textbf{Neutral} & \textbf{\begin{tabular}[c]{@{}c@{}}Not\\ Appropriate\end{tabular}} & \textbf{\begin{tabular}[c]{@{}c@{}}Not Appropriate\\ at all\end{tabular}}    \\ \hline
From original dataset          & 41\%                      & 54\%                 & 2\%               & 3\%                      & 0\%                             \\ \hline
{\begin{tabular}[c]{@{}c@{}}Sampled from\\ multi-reference collected\end{tabular}}
& 40\%                      & 52\%                 & 3\%               & 5\%                      & 0\%                             \\ \hline
\end{tabular}
\caption{Results from dataset quality experiment}
\label{table:dataquality}
\end{table*}

\section{Multi-Reference Data Collection}
We used the following procedure to prepare the DailyDialog test set for the multi-reference test set collection.
A dialogue \textit{D} in the test set consists of utterances $\{u_1, u_1,...,u_n\}$. 
Here, $u_i$ denotes the utterance at the $ith$ turn. For generating dialogue contexts, we truncate the dialogue at each possible utterance, except the last one. The response following each context is treated as the reference response. As an illustration, for the Dialogue shown in Table \ref{table:multirefeg}, we would generate the following context-reference pairs: \textit{Context 1:} ``911 emergency. What is the problem?'', \textit{Reference 1:} ``I would like to report a break-in.''. \textit{Context 2:} ``911 emergency ... report a break-in.'', \textit{Reference 2:} ``'When was this break-in?'.
In our multi-reference dataset, we expand each single-reference to a set of multiple references.



\subsection{Data collection Procedure}
We designed an interface for multi-reference data collection using 
Amazon Mechanical Turk (AMT). For every HIT, we asked an AMT worker to generate 4 diverse follow-up responses for a conversation. A snapshot of the data collection interface is shown in Figure \ref{fig:interface1} (Appendix). 
We provided instructions and examples to further clarify the task.
To maintain quality post data collection, we filter out responses collected from workers who either generated very short responses or entered the responses in very short amount of time consistently. 

\subsection{Data Quality}
Using the method described above, we collected 4 diverse responses for the 1000 dialogues in the test set, which consists of \textit{6740 contexts}.
To validate the quality of the collected dataset, an experiment on AMT is carried out for 100 contexts sampled randomly from the dataset. Workers are shown a dialogue context followed by 3 responses shuffled in a random order - 1) the original response from the dataset 2) a random response from the collected multi-references, and 3) a distractor response, irrelevant to the dialogue context. We use distractor responses to filter out poor annotations where the annotator gave high ratings to the distractor response.
We ask the workers to rate each of the 3 responses for a dialogue context on a scale of 1-5 for appropriateness, where 1 indicates \textit{Not Appropriate at all} and 5 indicates \textit{Very Appropriate}. We present the ratings from the experiment in Table \ref{table:dataquality} for the original responses from the dataset, and the responses from the multi-reference set. We observe that 92\% sampled responses from the multi-reference set are marked Appropriate or Very Appropriate.
Moreover, only 8\% of the responses are marked Not Appropriate or lower, compared to 5\% for the original reference set. This indicates that the collected reference set is close to the original reference set in quality. Furthermore, the responses are generated specifically for each context, they are coherent with the context.

%% file: experiments.tex
\section{Experiments}

This section describes the experiments we conducted to explore the effectiveness of multi-reference evaluation.


\begin{table*}[tb]
\small
\centering
\begin{tabular}{|c|c|c|c|c|c|c|c|c|}
\hline
\textbf{}        & \multicolumn{4}{c|}{\textbf{Single-reference}}                             & \multicolumn{4}{c|}{\textbf{Multiple-reference}}                           \\ \hline
\textbf{Metrics} & \textbf{Spearman} & \textbf{p-value} & \textbf{Pearson} & \textbf{p-value} & \textbf{Spearman} & \textbf{p-value} & \textbf{Pearson} & \textbf{p-value} \\ \hline
BLEU-1           & 0.0241            & 0.591            & 0.1183           & 0.008            & 0.1572            & 0.000            & 0.2190           & 0.000            \\ \hline
BLEU-2           & 0.0250            & 0.577            & 0.1803           & 0.000            & 0.2077            & 0.000            & 0.2910           & 0.000            \\ \hline
BLEU-3           & 0.0608            & 0.175            & 0.1269           & 0.005            & 0.2520            & 0.000            & 0.2086           & 0.000            \\ \hline
BLEU-4           & 0.0345            & 0.441            & 0.1380           & 0.002            & 0.2202            & 0.000            & 0.2333           & 0.000            \\ \hline
METEOR           & 0.1064            & 0.017            & 0.1871           & 0.000            & 0.2247            & 0.000            & 0.2855           & 0.000            \\ \hline
ROUGE-L          & 0.0715            & 0.110            & 0.1408           & 0.002            & 0.2203            & 0.000            & 0.2798           & 0.000            \\ \hline
Embedding Average          & 0.0301            & 0.502            & -0.0067          & 0.880            & 0.1248            & 0.005            & 0.0636           & 0.156            \\ \hline
Vector Extrema          & 0.1919            & 0.000            & 0.2114           & 0.000            & 0.2785            & 0.000            & 0.2946           & 0.000            \\ \hline
Greedy  Matching         & 0.1306            & 0.003            & 0.1150           & 0.010            & 0.2367            & 0.000            & 0.2352           & 0.000            \\ \hline
Skip-Thought    & -0.0029           & 0.949            & -0.1463          & 0.001            & 0.1049            & 0.019            & -0.0716          & 0.109            \\ \hline
GenSen           & 0.0731            & 0.103            & 0.1110           & 0.013            & 0.1832            & 0.000            & 0.2389           & 0.000            \\ \hline
\end{tabular}
\caption{Correlation of various metrics when evaluated using single-reference and multi-reference test sets. Evaluation using Multiple References leads to better correlation across all metrics.}
\label{table:correlationmetrics}
\end{table*}

\begin{figure*}[!htb]
\centering
  \begin{subfigure}[b]{0.45\textwidth}
  \centering
    \includegraphics[width=\textwidth]{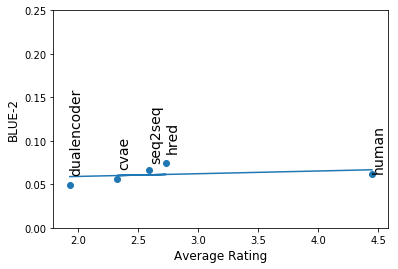}
    \caption{BLEU-2-human ratings - single-references}
    \label{fig:f1}
  \end{subfigure}
  \hfill
  \begin{subfigure}[b]{0.45\textwidth}
    \includegraphics[width=\textwidth]{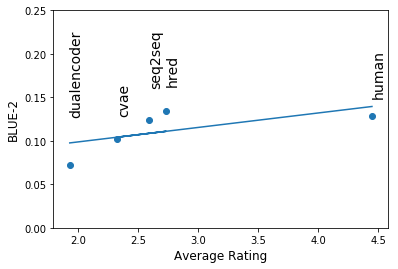}
    \caption{BLEU-2-human ratings - multiple references}
    \label{fig:f2}
  \end{subfigure}
  
    \begin{subfigure}[b]{0.45\textwidth}
    \includegraphics[width=\textwidth]{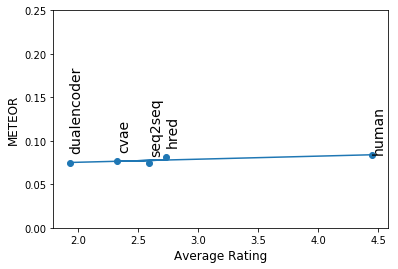}
    \caption{METEOR-human ratings - single-references}
    \label{fig:f3}
  \end{subfigure}
  \hfill
  \begin{subfigure}[b]{0.45\textwidth}
    \includegraphics[width=\textwidth]{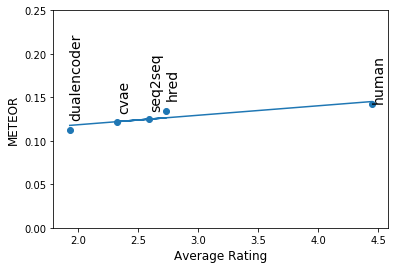}
    \caption{METEOR-human ratings - multiple references}
    \label{fig:f4}
  \end{subfigure}
  \caption{System level correlations for BLEU-2 and METEOR metrics. Multi-reference evaluation shows higher correlation with more clear differentiation in model performance.}
  \label{figure:systemcorr}
\end{figure*}

\subsection{Correlation Analysis for Quality}
This analysis aims to compute the correlation between human quality judgments and two forms of automatic evaluation, both single-reference and multi-reference.

\subsubsection{Human Annotations}
\label{section:correlation-data} A collection of 100 dialogue contexts are randomly selected from the dataset. For a particular dialogue context, each of the four models produces a response. In addition, we collect a human response using Amazon Mechanical Turk (AMT), making it total of five responses for each dialogue context. Given these context-response pairs, each response is rated in terms of appropriateness (from 1-5) by 5 different AMT workers. The ratings are removed for workers with a Cohen's Kappa $\kappa$ \cite{cohen1968weighted} inter-annotator agreement score of less than 0.2. The remaining workers had a mean $\kappa$ score of 0.43, indicating moderate agreement.

\subsubsection{Results}

\textbf{Utterance level correlation:}
The results of the correlation study conducted for 5 model responses for 100 contexts are shown in Table \ref{table:correlationmetrics}. Pearson correlation is computed to estimate linear correlation, and Spearman correlation to estimate monotonic correlation. The correlations with human quality judgments are computed for both single-reference and multi-reference evaluation. The multi-reference test set consists of both the original reference and the four new collected reference responses. For single-reference evaluation, except for METEOR and Vector Extrema metrics, the correlation is either small or statistically less significant. On the other hand, every metric shows higher and significant correlation for multi-reference evaluation, with METEOR, ROUGE-L and Vector Extrema achieving the highest correlation values. These results indicate that multi-reference evaluation correlates significantly better with human judgment than single-reference, across all the metrics. This reaffirms the hypothesis that multi-reference evaluation better captures the one-to-many nature of open-domain dialogue.



\vspace{2mm}
\noindent
\textbf{System level correlation:}
For each model used in the correlation study, the average human rating and average metric scores for 100 contexts are used to calculate system-level correlations. We show system-level correlations for metrics BLEU-2 and METEOR metrics in Figure \ref{figure:systemcorr}. Each point in the scatter plots represents the average scores for a dialogue model. Average human scores are shown on the horizontal axis, with average metric scores on the vertical axis. Humans ratings are low for responses from the retrieval model, and higher for human responses and responses from HRED model. It is clear that the difference in scores for models when evaluated using single-references is not significant enough to compare the models, as the average metric scores have near zero or very weak correlation with average human ratings. This renders them insufficient for dialogue evaluation. However, with multi-reference evaluation, the correlation is higher and significant, which differentiates the models clearly. Thus, multi-reference based evaluation correlates well with humans both at utterance level and at the system level.

\subsection{Correlation Analysis for Diversity}

This section aims to demonstrate that referenced diversity evaluation methods better correlate with \textit{human judgements of diversity}, than previously used unreferenced diversity metrics. While unreferenced metrics simply reward lexical differences amongst generated outputs, referenced methods (e.g., the recall metric) aims to calculate the \textit{coverage} of the responses. 
The correlation of human diversity scores is calculated with both unreferenced and referenced measures of diversity.

\subsubsection{Human Annotations}
Multiple hypotheses were generated from all the models. For CVAE, multiple responses are sampled from the latent space with greedy word-level decoding. For rest of the generation models, five responses were obtained using sampled decoding. For retrieval models, the top five retrieved responses were used. Human annotations of these multiple hypotheses were collected as follows: (1) Workers mark the responses which they find to be appropriate for the conversational context, (2) They then provide a score for the diversity of the responses based on how different they are in \textit{meaning}. This two-stage annotation process captures a desired form of system diversity: generated outputs should be varied, but also appropriate. 
The scores are averaged across the three workers' annotations. We filtered out ratings from workers with low inter-annotator agreement as described in section \ref{section:correlation-data}. The final mean $\kappa$ score of 0.41, which indicates moderate agreement.

\begin{table}[tb]
\small
\begin{tabular}{|@{\hskip4pt}c@{\hskip4pt}|@{\hskip4pt}c@{\hskip4pt}|@{\hskip4pt}c@{\hskip4pt}|@{\hskip4pt}c@{\hskip4pt}|@{\hskip4pt}c@{\hskip4pt}|}
\hline
\textbf{Metric}                                         & \textbf{\begin{tabular}[c]{@{}l@{}}Spearman\\ \end{tabular}} & \textbf{p-value} & \textbf{\begin{tabular}[c]{@{}l@{}}Pearson\\ \end{tabular}} & \textbf{p-value} \\ \hline
Distinct-1                                              & 0.0204                                                           & 0.647         & 0.0465                                                            & 0.299        \\ \hline
Distinct-2                                              & -0.1282                                                              & 0.004       & -0.0568                                                             & 0.205         \\ \hline
Distinct-3                                              & -0.1316                                                           & 0.003        & -0.0184 &    0.681      \\ \hline
\begin{tabular}[c]{@{}c@{}}Self\\ BLEU-2\end{tabular}                                          & -0.1534                                                              & 0.001        & -0.1251                                                            & 0.005         \\ \hline
\begin{tabular}[c]{@{}c@{}}Self\\ BLEU-4\end{tabular}                                                 & -0.0836                                                              & 0.061        & -0.0304                                                           & 0.497      \\ \hline
\begin{tabular}[c]{@{}c@{}}Recall\\ BLEU-2\end{tabular} & 0.2052                                                               & 0.000           & 0.2469                                                                 & 0.000          \\ \hline
\begin{tabular}[c]{@{}c@{}}Recall\\ BLEU-4\end{tabular} & 0.1713                                                                 & 0.000           & 0.1231                                                                & 0.005          \\ \hline

\begin{tabular}[c]{@{}c@{}}Recall\\ METEOR\end{tabular} & 0.1993                                                              & 0.000           & 0.2165                                                                & 0.000          \\ \hline
\begin{tabular}[c]{@{}c@{}}Recall\\ ROUGE-L\end{tabular} & 0.1862                                                                 & 0.000           & 0.2234                                                                & 0.000         \\ \hline

\begin{tabular}[c]{@{}c@{}}Recall \\Vector\\  Extrema\end{tabular} & 0.2063                                                                & 0.000           & 0.2314                                                              & 0.000         \\ \hline

\begin{tabular}[c]{@{}c@{}}Recall\\ Greedy\\Matching \end{tabular} & 0.0797                                                                 & 0.075          & 0.1204                                                               & 0.007         \\ \hline

\end{tabular}
\caption{Correlation scores for diversity metrics}
\label{diversitycorr}
\end{table}

\subsubsection{Results}
The results for the diversity correlation analysis are shown in Table \ref{diversitycorr} for a selected set of metrics\footnote{For Self-BLEU we calculate correlation with values substracted from 1 as Self-BLEU is inversely related to diversity}. The unreferenced metrics, Distinct and Self-BLEU, correlate poorly with human judgment. This is probably because these metrics evaluate lexical diversity, while humans evaluate diversity of meaning. Furthermore, unreferenced metrics do not consider the reference response and reward diverse outputs without considering appropriateness. 
With referenced diversity evaluation, using the recall method, BLEU-2 and Vector Extrema show the highest correlation. While metrics like Self-BLEU and Distinct can be ``gamed'' by producing meaningless albeit very diverse responses, the referenced recall metrics require both appropriate and diverse outputs. As such, referenced evaluation correlates significantly better with human notions of diversity. Thus, the construction of a multi-reference dataset allows for improved diversity metrics. 

\begin{table*}[]
\small
\centering
\begin{tabular}{|@{\hskip4pt}c@{\hskip4pt}|@{\hskip4pt}c@{\hskip4pt}|@{\hskip4pt}c@{\hskip4pt}|@{\hskip4pt}c@{\hskip4pt}|@{\hskip4pt}c@{\hskip4pt}|@{\hskip4pt}c@{\hskip4pt}|@{\hskip4pt}c@{\hskip4pt}|@{\hskip4pt}c@{\hskip4pt}|@{\hskip4pt}c@{\hskip4pt}|@{\hskip4pt}c@{\hskip4pt}|@{\hskip4pt}c@{\hskip4pt}|}
\hline
\textbf{}                                                 & \multicolumn{5}{c|}{\textbf{Single-reference}}                                                                                      & \multicolumn{5}{c|}{\textbf{Multiple-reference}}                                                                                    \\ \hline
\textbf{Metric}                                           & \textbf{\begin{tabular}[c]{@{}c@{}}Dual\\ Encoder\end{tabular}} & \textbf{Seq2Seq} & \textbf{HRED} & \textbf{CVAE} & \textbf{Human} & \textbf{\begin{tabular}[c]{@{}c@{}}Dual\\ Encoder\end{tabular}} & \textbf{Seq2Seq} & \textbf{HRED} & \textbf{CVAE} & \textbf{Human} \\ \hline
BLEU-2                                                    & 0.0399                                                          & 0.0521           & 0.0604        & 0.0656        & 0.0513         & 0.0625                                                          & 0.0981           & 0.1061        & 0.1033        & 0.1637         \\ \hline
BLEU-4                                                    & 0.0168                                                          & 0.0252           & 0.0301        & 0.0291        & 0.0245         & 0.0241                                                          & 0.0445           & 0.0497        & 0.0429        & 0.0791         \\ \hline
METEOR                                                    & 0.0653                                                          & 0.0544           & 0.0607        & 0.0724        & 0.0592         & 0.1000                                                          & 0.0970           & 0.1036        & 0.1120        & 0.1456         \\ \hline
ROUGE-L                                                   & 0.1522                                                          & 0.1847           & 0.1998        & 0.2088        & 0.1682         & 0.2216                                                          & 0.2927           & 0.3044        & 0.2997        & 0.3502         \\ \hline  
\begin{tabular}[c]{@{}c@{}}Vector\\ Extrema\end{tabular}  & 0.4005                                                          & 0.5124           & 0.5002        & 0.4893        & 0.4823         & 0.4713                                                          & 0.6191           & 0.5975        & 0.5722        & 0.6134         \\ \hline
\begin{tabular}[c]{@{}c@{}}Greedy\\ Matching\end{tabular} & 0.6257                                                          & 0.7167           & 0.7104        & 0.7078        & 0.6799         & 0.6991                                                          & 0.7649           & 0.7551        & 0.7457        & 0.7562         \\ \hline \hline
\begin{tabular}[c]{@{}c@{}}Recall\\ BLEU-2\end{tabular}           & 0.0662                                                          & 0.0544           & 0.0766        & 0.1077        & 0.0898         & 0.0436                                                          & 0.0377           & 0.0556        & 0.0679        & 0.0984         \\ \hline
\begin{tabular}[c]{@{}c@{}}Recall\\ Vector\\ Extrema\end{tabular} & 0.4945                                                          & 0.5127           & 0.5397        & 0.5586        & 0.5651         & 0.4934                                                          & 0.5334           & 0.5476        & 0.5653        & 0.5881         \\ \hline

\end{tabular}
    \caption{Model evaluation with automatic metrics on Single and Multiple references. Multiple reference evaluation is able to correctly rank human responses higher than model responses.}
    \label{table:modelhuman}
\end{table*}

\subsection{Automatic Evaluation of Models}
\label{section:modelhu}
We use our multi-reference evaluation methodology to compare the models and the human generated responses on the whole test dataset. For the human model, we use one reference from the multi-reference set as the hypothesis. 
Human responses are generally more interesting and diverse than model responses, which are known to suffer from the dull response problem \citep{li2016deep}. Because of this reason, we would expect the human generated responses to get higher scores than the  dialogue models. However, the results presented in Table \ref{table:modelhuman} show that single-reference automatic evaluation ranks few models higher than the humans model. With multi-reference evaluation, human performance is significantly higher than model performance. We further present scores for diversity metrics on multiple hypothesis generated for 100 contexts in the last two rows of the table. The use of multi-reference evaluation covers a wider array of valid responses, which strongly rewards the diverse human responses compared to single-reference evaluation. 


\begin{figure}
    \centering
    \includegraphics[width=0.5\textwidth]{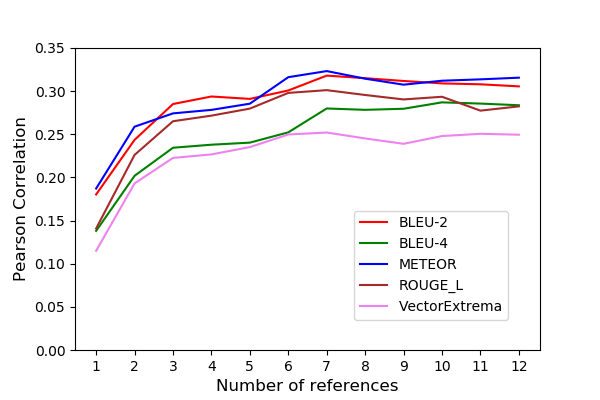}
    \caption{Change in correlation with varying number of references. Trend stablizes after 4-5 references}
    \label{figure:refmetrics}
\end{figure}

\begin{table}[tb]
\small
    \centering
    \begin{tabular}{l}
    \hline
\textbf{Dialogue Context:}\\
\textit{Person A:} excuse me . check please .\\
\hline 
\textbf{Generated Response} \\
sure , i 'll grab it and be right with you .
\\
\hline
\textbf{Single-reference Response:} \\
ok , how was everything ?
\\
\hline
\textbf{Multiple-reference Responses:}\\
i 'll get it right away .\\
here is the check .\\	no problem , let me get your server .\\
i 'll be right back with it .\\
\hline
Average Human Rating: 5\\
\begin{tabular}{@{\hskip4pt}c@{\hskip4pt}@{\hskip4pt}|@{\hskip4pt}c@{\hskip4pt}|@{\hskip4pt}c}
\hline
Metric & Single reference & Multiple reference\\
\hline
BLEU-2 & 0.0275 &0.3257 \\
METEOR  & 0.0539& 0.3425\\
Vector Extrema  & 0.5523&0.8680\\
\hline
\end{tabular}

\end{tabular}
\caption{Example of difference in metric scoring for single versus multiple reference evaluation.}
\label{table:quality}
\end{table}

\subsection{Effect of number of references}
The correlation of automated evaluation with human judgment is calculated at various numbers of reference responses. The results shown in Figure \ref{figure:refmetrics} demonstrate that the Pearson correlation with human judgment generally increases sharply up to 3-5 references. It further increases slowly up to about 7 references and then seems to plateau at around eight references. This suggests that four to eight references give sufficient coverage of the response space, and collecting additional references does not provide much value in terms of mitigating the issues of the one-to-many problem.

\section{Discussion and Conclusion}
This work proposes a more reliable methodology for automatic evaluation of open-domain dialogues with the use of multiple references. We augment the test set of DailyDialog dataset with multiple references and show that multiple references lead to better correlation with human judgments of quality and diversity of responses. Single-reference based evaluation can unfairly penalize diverse and interesting responses which are appropriate, but do not match a particular reference in the dataset. However, multiple references can cover the possible semantic space of replies for a context better than a single reference. Thus using multi-reference test sets can improve the way open-ended dialogue systems are currently evaluated. Our experiments also show that human-generated responses perform worse than models across most metrics when using single-reference evaluation, but multiple reference evaluation consistently ranks human responses higher than model-generated responses. Furthermore, we show how varying the number of references effects human judgement correlation. This methodology could easily be extended to other open domain datasets 
if the community can make similar multi-reference test sets publicly available. 

We illustrate the strength of multi-reference evaluation through scores calculated for some metrics using both single and multiple references for an example context in Table \ref{table:quality}. Multiple reference-based evaluation is often good at assigning higher scores when there is more scope for diversity in the responses as illustrated by the example. It should be noted that multiple reference evaluation generally increases the scale of metrics for all responses, and this includes dull responses.

The multi-reference data collection procedure in this paper collects the same number of responses for all contexts. However, different dialogue contexts might possess different levels of ``open-endedness''. For e.g., a context like ``Would you like to dance?'' would generally have fewer possible variations in responses than a more open-ended context like ``What did you do yesterday?''. Therefore, the number of references to collect for a context could be based on the expected variability in responses for the context. Such a procedure would capture more variability over the dataset for a fixed budget.

An important direction in dialogue system research is to build models that have more engaging and meaningful conversations with a human. With the recent push towards models which can generate more diverse and interesting responses, appropriate evaluation methodologies are an important and urgent need for the community. Human level evaluation of generation and diversity is challenging to do in a completely automatic way, however, compared to evaluating with a single response, we show that the proposed evaluation methodology is more reliable and will facilitate progress in this direction. In this work we have chose one dataset for extensive experimentation, but in the future studies, it will be worth collecting more datasets and repeating the correlation experiments.

%% file: acknowledgements.tex
\section{Acknowledgements}
This work was funded by the Defense Advanced Research Planning Agency (DARPA) under DARPA Grant N66001–98-1–8908, and the National Science Foundation under Award \#IIS-1816012. We thank the workers on Amazon Mechanical Turk for making our research possible.